# Fully Automatic Electrocardiogram Classification System based on Generative Adversarial Network with Auxiliary Classifier

**Zhanhong Zhou[1], Xiaolong Zhai[1], Chung Tin[1]**

## Abstract

A generative adversarial network (GAN) based fully automatic electrocardiogram (ECG) arrhythmia classification system with high performance is presented in this paper. The generator ($G$) in our GAN is designed to generate various coupling matrix inputs conditioned on different arrhythmia classes for data augmentation. Our designed discriminator ($D$) is trained on both real and generated ECG coupling matrix inputs, and is extracted as an arrhythmia classifier upon completion of training for our GAN. After fine-tuning the $D$ by including patient-specific normal beats estimated using an unsupervised algorithm, and generated abnormal beats by $G$ that are usually rare to obtain, our fully automatic system showed superior overall classification performance for both supraventricular ectopic beats (SVEB or S beats) and ventricular ectopic beats (VEB or V beats) on the MIT-BIH arrhythmia database. It surpassed several state-of-art automatic classifiers and can perform on similar levels as some expert-assisted methods. In particular, the $F_1$ score of SVEB has been improved by up to 10% over the top-performing automatic systems. Moreover, high sensitivity for both SVEB (87%) and VEB (93%) detection has been achieved, which is of great value for practical diagnosis. We, therefore, suggest our ACE-GAN (Generative Adversarial Network with Auxiliary Classifier for Electrocardiogram) based automatic system can be a promising and reliable tool for high throughput clinical screening practice, without any need of manual intervene or expert assisted labeling.

## 1. Introduction

Electrocardiogram (ECG) is a clinical standard for diagnosing heart-related diseases, and is particularly valuable for arrhythmia screening and classification. Nevertheless, the interpretation of ECG signal requires extra expertise. Given the large amount of ECG recordings collected in daily clinical routine, interpreting ECG by cardiologists manually is not only resource-exhaustive but also time consuming. Hence, fully automatic heartbeats classification systems with reliable performance are highly recommended. Various classification systems have been proposed using techniques based on discrete wavelet transform (Ye, Kumar, & Coimbra, 2012), time-domain feature extraction (Mazomenos et al., 2012), frequency-domain feature extraction (Romero & Serrano, 2001), feature selection methods (Llamedo & Martinez, 2011; Mar, Zaunseder, Martínez, Llamedo, & Poll, 2011) and abstract feature (Teijeiro, Félix, Presedo, & Castro, 2018), etc. Recently, data-driven techniques including convolutional neural network (CNN) which can extract features automatically have started to attract a lot of interests. CNN has shown its strength in various machine learning tasks, including biosignal classification (Ronneberger, Fischer, & Brox, 2015; Zhai, Jelfs, Chan, & Tin, 2017) and image recognition (Krizhevsky, Sutskever, & Hinton, 2012; Simonyan & Zisserman, 2015). Moreover, it is revealed by previous studies (Acharya et al., 2017; Hannun et al., 2019; Kiranyaz, Ince, & Gabbouj, 2016; Zhai & Tin, 2018; Zhai, Zhou, & Tin, 2020) that typical CNN based systems can provide superior performance in arrhythmia classification without the need of hand-crafted feature extraction.

[1]Department of Biomedical Engineering, City University of Hong Kong, Hong Kong.
Correspondence to: Tin Chung <chungtin@cityu.edu.hk>

Despite the advantages of typical CNNs in arrhythmia diagnosis tasks, they also face several challenges. First, arrhythmia ECG are commonly of imbalanced classes, where normal beats (N beats) outnumber abnormal beats such as supraventricular ectopic beats (SVEB or S beats) and ventricular ectopic beats (VEB or V beats) (A. L. Goldberger et al., 2000; Hermes, Geselowitz, & Oliver, 1980; Moody & Mark, 2001). For example, 90% of the beats in the MIT-BIH arrhythmia database are normal beats. Some workarounds have been proposed to resolve such a class imbalance problem, including cost-sensitive learning and random re-sampling (Chawla, Japkowicz, & Kotcz, 2004). However, both random over-sampling and cost-sensitive learning such as assigning different weights to each class (Xu, Mak, & Cheung, 2018) could overfit the training data and possibly reduce generalization in actual application. On the other hand, random under-sampling, where a subset of beats can be selected from the overall dataset to fulfill balanced classes (Kiranyaz et al., 2016; Zhai & Tin, 2018), would potentially discard important samples and the training set could become too small for training CNN effectively.

Second, the limited availability of labeled ECG data (especially for arrhythmic classes) restricts the effectiveness of supervised learning for training these classifiers because overfitting would likely to happen. Moreover, as physiological signals suffer from significant inter-subject variability, these CNN classifiers might perform poorly in un-seen patients. As such, previous studies (Chazal, 2014; Chazal & Reilly, 2006; Llamedo & Martinez, 2012; Rahhal et al., 2016; Xia et al., 2018; Ye, Kumar, & Coimbra, 2016) have proposed to include some labeled patient-specific heartbeats for boosted classification performance. To some extent, this could also alleviate the class imbalance problem, since patient-specific arrhythmic beats could be annotated to improve the performance of patient-dependent classifiers. However, assigning annotations patient by patient would be extremely costly and consequently, those systems are not fully automatic (require expert assistance).

To tackle the challenges above and meanwhile exploit the benefits of CNN, we suggest that the generative adversarial network (GAN) (Goodfellow et al., 2014) is a promising approach to overcome these. GAN usually contains two neural networks, the generator ($G$) and the discriminator ($D$), that can play a zero-sum game by competing and cooperating with each other. In brief, $G$ learns to generate "fake" data that looks like the real one to deceive $D$; while $D$ learns to discriminate the fake data generated by $G$ from the real data. As such, GAN has shown great success in generating images of specific types from noise (Mirza & Osindero, 2014; Reed et al., 2016), transferring images styles (Chen et al., 2016; Isola, Zhu, Zhou, & Efros, 2017; Zhu, Park, Isola, & Efros, 2017), interpolating high-resolution images (Ledig et al., 2017; Odena, Olah, & Shlens, 2017), and so on. Some studies have also reported the success of GAN in biomedical applications, for example, to de-noise CT images (Wolterink, Leiner, Viergever, & Išgum, 2017) and stain tissue-autofluorescence images (Rivenson et al., 2019). For ECG arrhythmia classification, the GAN may provide data augmentation and relieve class imbalance problem by learning to generate relevant conditional samples. On the other hand, some researchers have looked into the trained $D$ from GAN and shown improved performance using it as classifier than traditional deep neural network (DNN) (Odena, 2016; Shen, Lu, Li, & Kawai, 2017; Springenberg, 2016).

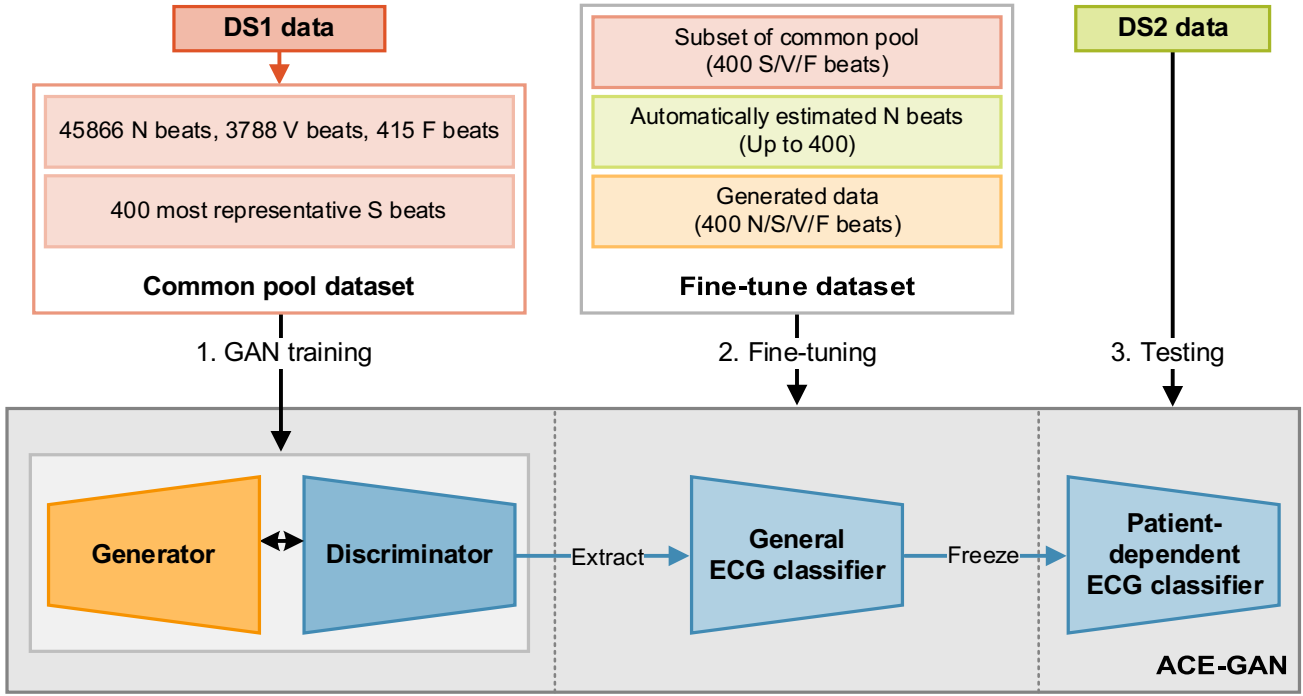

**Fig. 1.** Schematic of the proposed ACE-GAN based fully automatic arrhythmia classification system. The ACE-GAN is firstly trained with the common pool dataset from DS1. The ECG classifier is extracted from the trained discriminator and is fine-tuned. The fine-tune dataset comprises of the subset of common pool dataset, the automatically estimated normal beats from the patient, and the generated data from the trained generator as data augmentation. The ECG classifier then serves as the final classifier to predict heartbeat labels for records in DS2.

In this work, we adopted the GAN framework and have designed the ACE-GAN (Generative Adversarial Network with Auxiliary Classifier for ECG). This system includes the $G$ for data augmentation, and extracts the well-trained $D$ as the classifier. Our system is fully automatic and does not require any manual patient-specific labeling. We tested it on the MIT-BIH arrhythmia database (A. L. Goldberger et al., 2000; Moody & Mark, 2001) and compared its performance against previous studies with or without expert assistance following AAMI recommendation (ECAR, 1987). The overall workflow of the proposed systems is summarized in Fig. 1.

## 2. Related work

Recently, the advantages of CNN for automatic feature extraction and classification in ECG arrhythmia time series signals have been explored. Instead of designing specific features manually with expert knowledge, (Kiranyaz et al., 2016) proposed a 1-D CNN for accurate arrhythmia detection of SVEB and VEB. The classifier was trained with common pool data and subject-specific training data patient by patient and had provided top-tier performance at that time. Later, a 2-D CNN system for arrhythmia classification was reported by (Zhai & Tin, 2018). By integrating adjacent heartbeats into image-like 2D inputs, the proposed CNN achieves over 10% increment in SVEB detection against top ranking algorithms, while maintaining comparable VEB detection performance. In the work of (F. Li, Xu, Chen, & Liu, 2019), they presented a 3-D CNN inputs based arrhythmia classification system for deep features extraction. These studies have shown promising use of CNN in SVEB and VEB detection. More deep learning methods in arrhythmia classification is reviewed in (Ebrahimi, Loni, Daneshtalab, & Gharehbaghi, 2020).

Nevertheless, these methods require manual annotations of a portion of the testing subjects' heartbeats (e.g. 5 min. of heartbeat for each record as common practice) to realize a patient-specific classifier (Kiranyaz et al., 2016; F. Li et al., 2019; Zhai & Tin, 2018). The classification performance would be compromised otherwise due to inter-subject variability. In order to minimize the work of further manual annotation, (Rahhal et al., 2016) and (Xia et al., 2018) proposed to select only the most uncertain or the most informative beats in the testing subjects to be annotated by medical experts during the course of training of DNNs. Likewise, (Llamedo & Martinez, 2012) described an arrhythmia classifier with expert assistance, where they showed that manual annotation of up to 12 beats per testing subject is required to yield good performance. On the other hand, to realize a fully automatic system, (Ye et al., 2016) proposed a subject-adaptable ECG classifier based on a multi-view learning framework. A few high-confidence beats were selected from the testing subject's record, based on labels given by preliminary classifiers that trained with common set data. They also selected some training samples with similar N beat patterns as the testing subject's, and 50-150 representative S or V beats to be included in the training data. Recently, a deep learning based automatic approach for inter-subject ECG classification was reported in (Niu, Tang, Sun, & Zhang, 2020), where CNN was used to extract features from their symbolized heartbeat signals.

In view of these, our proposed model adopted the GAN approach to offer both data augmentation and effective classification automatically in a unified framework. We evaluated our system against several state-of-art methods. These methods, and ours, have all been developed and evaluated on the MIT-BIH arrhythmia database and followed the same data handling method, i.e. they all separated the database into DS1 and DS2 as in (Chazal, O'Dwyer, & Reilly, 2004), and evaluated subject-by-subject. We made comparison on sensitivity and positive predictive rate in SVEB and VEB detection, in particular.

## 3. Data preparation

### 3.1. Database

To evaluate the performance of the proposed ECG classification systems in this study, we used the publicly accessible MIT-BIH arrhythmia database (A. L. Goldberger et al., 2000; Moody & Mark, 2001). This database has been used in numerous previous studies (Chazal et al., 2004; Chazal & Reilly, 2006; Hu, Palreddy, & Tompkins, 1997; Jiang & Kong, 2007; Kiranyaz et al., 2016; Llamedo & Martinez, 2011, 2012; Mar et al., 2011; Rahhal et al., 2016; Raj & Ray, 2018; Teijeiro et al., 2018; Xu et al., 2018; Ye et al., 2012; Ye et al., 2016; Zhai & Tin, 2018; Zhang, Dong, Luo, Choi, & Wu, 2014) and serves as the golden standard for evaluating various arrhythmia classifiers. In this database, 48 ECG recordings from 47 patients are included. Each recording contains two-channel ECG data sampled at 360 Hz for 30 min. Each heartbeat in the record was annotated by at least two cardiologists independently. As defined by AAMI (ECAR, 1987), heartbeats in this database can be classified into five types, including N (beats originating in the sinus node), S (supraventricular ectopic beats), V (ventricular ectopic beats), F (fusion beats) and Q (unclassifiable beats). The detailed mapping between MIT-BIH arrhythmia database heartbeat classes and AAMI standard classes is shown in Table 1. Four recordings (#102, #104, #107 and #217) that contain paced beats were excluded in this study.

The records in the MIT-BIH database was divided into DS1 and DS2 (Table 2), as in many previous studies (Chazal et al., 2004; Chazal & Reilly, 2006; Llamedo & Martinez, 2011, 2012; Mar et al., 2011; Raj & Ray, 2018; Teijeiro et al., 2018; Ye et al., 2016; Zhang et al., 2014). DS1 and DS2 each contain 22 ECG records respectively. DS1 was used as training set while DS2 was regarded as testing set.

**Table 1**
Mapping between MIT-BIH arrhythmia database classes and AAMI classes.

| AAMI | | MIT-BIH Arrhythmia Database | |
|---|---|---|---|
| Classes | Beat Types | Classes | Beat Types |
| N | Beats originating in the sinus node | · or N | Normal beat |
| | | L | Left bundle branch block beat |
| | | R | Right bundle branch block beat |
| | | e | Atrial escape beat |
| | | j | Nodal (junctional) escape beat |
| S | Supraventricular ectopic beats | A | Atrial premature beat |
| | | a | Aberrated atrial premature beat |
| | | J | Nodal (junctional) premature beat |
| | | S | Supraventricular premature beat |
| V | Ventricular ectopic beats | V | Premature ventricular contraction |
| | | E | Ventricular escape beat |
| F | Fusion beats | F | Fusion of ventricular and normal beat |
| Q | Unclassifiable beats | / | Paced beat |
| | | f | Fusion of paced and normal beat |
| | | Q | Unclassifiable beat |

**Table 2**
Record number and beats number in DS1 and DS2.

| | DS1 | DS2 | Total |
|---|---|---|---|
| Record # | 101, 106, 108, 109, 112, 114, 115, 116, 118, 119, 122, 124, 201, 203, 205, 207, 208, 209, 215, 220, 223, 230 | 100, 103, 105, 111, 113, 117, 121, 123, 200, 202, 210, 212, 213, 214, 219, 221, 222, 228, 231, 232, 233, 234 | - |
| N | 45866 | 44259 | 90125 |
| S | 944 | 1837 | 2781 |
| V | 3788 | 3221 | 7009 |
| F | 415 | 388 | 803 |
| Q | 8 | 7 | 15 |
| Total | 51021 | 49712 | 100733 |

## 3.2. Dual-beat coupling matrix input for the classification system

In this study, the modified lead II channel signal was used. The time series ECG signal from the database was preprocessed in a similar way as our previous work (Zhai & Tin, 2018). The procedure is briefly repeated here.

For each record in both DS1 and DS2, the segmentation length of one heartbeat is defined as the R-R interval averaged over the whole 30-minute record. Each beat segment is centered on its own R peak. The timing of R peak for each heartbeat is obtained directly from the database, which could, on the other hand, be robustly detected by numerous well-developed methods automatically (Bote, Recas, Rincón, Atienza, & Hermida, 2018; Jimenez-Perez, Alcaine, & Camara, 2020; C. Li, Zheng, & Tai, 1995; Martinez, Almeida, Olmos, Rocha, & Laguna, 2004; Pan & Tompkins, 1985; Peimankar & Puthusserypady, 2021; J. Wang, Li, Li, & Fu, 2020). Development of new R peak detection algorithm is beyond the focus of this study. To compute the dual-beat coupling matrix, two pairs of adjacent heartbeats are considered together. The first pair consists of the current beat and the previous beat, denoted as the column vector:

$$Dual\_beat_{i-1,i} = Beat_{i-1}[1] \cdots Beat_{i-1}[k] \cdots Beat_{i-1}[L], \ Beat_i[1] \cdots Beat_i[k] \cdots Beat_i[L] \qquad (1)$$

where $Beat_i[k]$ is the $k^{th}$ sample point of the current beat in time, and $L$ is the original length of beats in this record. The second pair consists of the current beat and the next beat (denoted as the column vector $Dual\_beat_{i,i+1}$).

These two dual-beat vectors are then scaled respectively to the required input size $M$ as follows. First, the original vector with length $2L$ is interpolated by a factor equals to the input size $M$. Then, the average values for every $2L$ data points are calculated. As a result, a vector of length $M$ is obtained from any original length of $2L$.

After that, we compute the coupling matrix (CM) with size $M \times M$, as follows,

$$\text{CM} = \begin{bmatrix} Dual_{beat_{i-1,i}}^{scaled}[1] \\ \vdots \\ Dual_{beat_{i-1,i}}^{scaled}[M] \end{bmatrix} \cdot \begin{bmatrix} Dual_{beat_{i,i+1}}^{scaled}[1] & \cdots & Dual_{beat_{i,i+1}}^{scaled}[M] \end{bmatrix} \tag{2}$$

The input size $M$ is defined as 73 following our previous work (Zhai & Tin, 2018). Our classification system is designed to take this coupling matrix as input and provide a label for the center beat, $Beat_i$.

### 3.3. Common pool dataset

As shown in Fig. 1, our ACE-GAN is first trained using the common pool dataset, which includes 50469 beats (all 45866 N beats, all 3788 V beats, all 415 F beats, and 400 most representative S beats) in DS1. In our previous work (Zhai & Tin, 2018), we have shown that the S beats should be systematically selected for the training set to achieve good classification performance. Furthermore, around half of the S beats in DS1 (~400 beats) came from a single subject. If the S beats were selected randomly for our training, the classification performance is bound to be compromised. The selection of S beats follows the method in (Zhai & Tin, 2018). Briefly, a binary classifier (with the same structure as $D$) is constructed to discriminate N and S beats. Each time, 75 N and 75 S beats are randomly selected from 16 records of DS1 (where S beats exist) to train this classifier. This classifier is then tested on 100 N beats randomly selected from the remaining 6 records in DS1. This was repeated 200 times using different sets of 75 S beats. After that, we select the 400 S beats in DS1 that give the highest average prediction accuracy for N beats in this procedure. These 400 most representative S beats selected are expected to have significantly different morphologies from the N beats and hence should be more useful for the subsequent training (Zhai & Tin, 2018). We kept 400 S beats in order to maintain a balanced fine-tuning dataset as explained below. On the other hand, selection of V beats in this step is not necessary since most V beats have distinct waveforms than N or S beats (Ary L Goldberger, Goldberger, & Shvilkin, 2017). In fact, we observed that applying this method to select 400 V beats resulted in reduced classification performance at the end. It is preferable to use all of the V beats for training. Q beats are not used for training in this study, as these unclassifiable beats are unlikely to help boost classification performance.

### 3.4. Fine-tune dataset

Our fine-tune dataset (Fig. 1) comprises of three parts: i) the 400 most representative S beats as described above, along with 400 V beats and 400 F beats randomly selected from common pool dataset; ii) a new pool of N, S, V and F beats (400 for each class) generated by the well-trained $G$; and iii) up to 400 automatically estimated patient-specific N beats as introduced below. Since the ~400 F beats in DS1 represent the minority class, we used 400 real beats each from other classes to maintain a balanced fine-tune dataset. All these beats are assigned with a label of "real" beat.

### 3.5. Estimation of patient-specific normal beats

Since including patient-specific heartbeats for training is expected to improve performance (Chazal, 2014; Chazal & Reilly, 2006; Llamedo & Martinez, 2012; Rahhal et al., 2016; Xia et al., 2018; Ye et al., 2016), we here used a simple unsupervised method to extract N beats from the testing subject reliably (Zhai et al., 2020). Briefly, each time, we consider the same pair of dual-beats segments as in computing the coupling matrix. The spectrograms for both dual-beats segments are computed using short-time Fourier Transform (STFT) respectively. We use the 1024-point Fast Fourier Transform (FFT) with a window size of 64 and a step of 1. The spectrogram power is retained only up to 11.23Hz (corresponds to 32 points) where most power of ECG is found. Then, the two spectrograms are unrolled and their correlation coefficient is calculated. A high correlation coefficient (close to 1) generally implies local regularity of waveform and rhythm of these three adjacent beats and they are likely to be N beats. Based on this principle, we can use the following algorithm to estimate a number of N beats from each testing record in an unsupervised way:

i) For each $Beat_i$, two spectrograms are computed respectively. If the correlation coefficient between them is larger than 0.9, $Beat_i$ will be considered as a normal beat and is stored in the N beat pool ($Est\_N\_Beats$).

ii) If the correlation coefficient is lower than 0.9, the spectrograms will be compared with those calculated from the N beat pool ($Est\_N\_Beats$) one by one. If any of the resulting correlation coefficients is larger than a threshold, $Beat_i$ can still be considered as an N beat. The threshold is defined as 0.95+epoch/100. This increasingly stricter criterion helps reduce false positive as the N beat pool ($Est\_N\_Beats$) increases in size.

iii) This procedure is repeated until no more beat is added to the N beat pool ($Est\_N\_Beats$) or when the pool size has reached the predefined requirement.

The accuracy of this estimating method for each record in DS2 is shown in Section 7. We emphasize that this unsupervised estimation method does not require any manual labeling/intervention, and hence our arrhythmia classification system remains fully automatic. The computational load of this algorithm is minor compared with training of the neural network. For each beat, only one spectrogram is computed and in the worst case scenario, up to 400 simple correlation operations are required.

## 4. ACE-GAN

The most important component in our classification system is the ACE-GAN. Fig. 2 presents the architecture of our proposed $G$ and $D$.

### 4.1. Generator architecture

As shown in Fig. 2a, $G$ takes a scalar class label and a $100 \times 1$ vector of Gaussian noise with zero mean and standard deviation of one ($\sim \mathcal{N}(0,1)$) as inputs. The label can be any of the four values representing N, S, V or F beat for conditioned generation. This scalar is then converted into a $100 \times 1$ vector using the embedding function in Keras (Chollet, 2015). The elementwise multiplication of the noise and the embedded label is taken at the first hidden layer of $G$. This hidden layer further connects to two layers in parallel through full connection, which correspond to the two pairs of adjacent beats required to calculate the coupling matrix (see Section 3.2). Batch normalization is deployed before these two layers (Ioffe & Szegedy, 2015). Each of these two layers goes through another full connection operation to fit into an input size of 73, and are finally used to compute the coupling matrix as in Eq. 2. We used rectified linear activation units (ReLU) as activation functions for all hidden layers but not for the last hidden layer of $G$. The 73-sized layers of $G$ are modeled as linear layers such that the magnitudes of the output coupling matrix are not limited to ±1, in contrast to the commonly used tanh activation. We observed that amplitudes of ECG signal typically vary over a wide range, and this linear activation allows our $G$ to generate coupling matrices of different classes with higher fidelity. The values of the hyper-parameters (e.g. numbers of nodes for each hidden layer) for our $G$ did not have significant impact on the final classification performance. As such, without loss of generosity, we used 3 hidden layers with a maximum of 256 nodes for our $G$. Alternatively, an automatic architecture search (e.g.(Gong, Chang, Jiang, & Wang, 2019)) can be considered in the future that may further boost the performance.

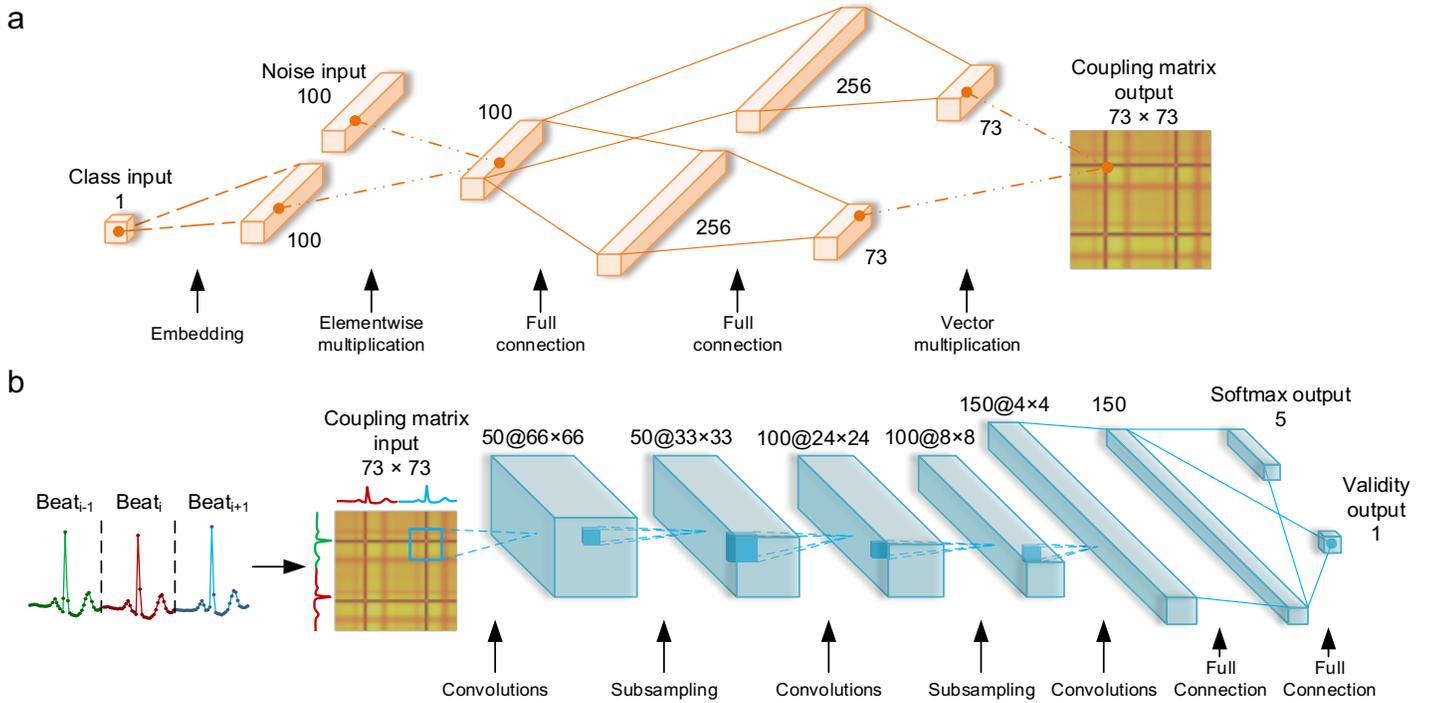

**Fig. 2.** Schematic of generator architecture (a) and discriminator architecture (b) in the proposed ACE-GAN. The generator takes a 1-dimensional scalar as the designated class label and a 100-dimensional noise vector as inputs. The label scalar is then embedded into a 100-dimensional vector to elementwise multiply with the noise vector input. The output of the generator is one coupling matrix computed by vector multiplication of two 73-dimensional vectors. The discriminator receives the 73 by 73 coupling matrix (computed from 2 pairs of adjacent heartbeat waveforms) as input while outputs a 1-dimensional scalar for validity and a 5-dimensional softmax layer indicating the predicted class label simultaneously. Details are described in Section 4.

### 4.2. Discriminator architecture

We herein used the CNN structure in (Zhai & Tin, 2018) as our $D$ in this study (Fig. 2b). Such CNN was originally adapted from the well-recognized LeNet (Lecun, Bottou, Bengio, & Haffner, 1998), and has shown promising performance in arrhythmia classification (Zhai & Tin, 2018). In brief, it comprises of a convolutional layer, a max pooling layer, a second convolutional layer, an average pooling layer and a third convolutional layer with kernel sizes of 8, 2, 10, 3 and 5, respectively. Then a fully connected layer is added in the latter part of this CNN. We again used ReLU as activation functions for these hidden layers. Dropout (Srivastava, Hinton, Krizhevsky, Sutskever, & Salakhutdinov, 2014) is applied to the last two hidden layers with a rate of 0.5. In this study, a linear unit is added in parallel to the softmax layer as output, which predicts whether the input comes from the real training data or is generated by $G$.

### 4.3. Objective functions and optimization

A variant of GAN named AC-GAN (GAN with auxiliary classifier) has been proposed to generate class-conditional samples (Odena et al., 2017). In AC-GAN, $G$ takes random noise input $z \sim p_z$ and class labels $c \sim p_c$ as inputs to generate $\mathcal{X}_{fake} = G(c, z)$. As a result, every generated sample corresponds to a class label and noise input. The $D$ in AC-GAN takes $\mathcal{X}_{fake}$ or $\mathcal{X}_{real} = x \sim p_{data}$ as input, and outputs the probability indicating the chance of the sample coming from a certain source $S$ (real or "fake"), namely $P(S \mid \mathcal{X})$; as well as the probability of belonging to a certain class $C$, namely $P(C \mid \mathcal{X})$ respectively. Therefore, two objective functions are considered: $\mathcal{L}_S$ for the likelihood of correct source and $\mathcal{L}_C$ for the likelihood of correct class. By training the AC-GAN, $D$ learns to maximize $(\mathcal{L}_C + \mathcal{L}_S)$ while $G$ learns to maximize $(\mathcal{L}_C - \mathcal{L}_S)$.

$$\mathcal{L}_S = E[\log P(S = s_{real} \mid \mathcal{X}_{real})] + E[\log P(S = s_{fake} \mid \mathcal{X}_{fake})] \tag{3}$$

$$\mathcal{L}_C = E[\log P(C = c \mid \mathcal{X}_{real})] + E[\log P(C = c \mid \mathcal{X}_{fake})] \tag{4}$$

In this work, we designed the ACE-GAN based on AC-GAN to fit in the task of ECG arrhythmia classification. Briefly, $G$ also takes noise $z \sim p_z$ and heartbeat class labels $c \sim p_c$ (i.e. N, S, V or F) as inputs to generate designated dual-beat coupling matrix samples, $\mathcal{X}_{fake}$. $D$ is trained on both real heartbeat coupling matrices $\mathcal{X}_{real}$ and the generated ones $\mathcal{X}_{fake}$, in order to predict the validity $S$ as well as the heartbeat class $C$ that $\mathcal{X}$ represents.

During the optimization of $D$, a real coupling matrix input is expected to receive validity of 1 at the linear unit (and 0 for a generated input) as well as be annotated with a class label indicating the type of beat (N, S, V, F or generated) from the softmax layer. The additional "generated" class label in the softmax layer is included to ensure that $D$ learns to classify beat types only from "real data". We found that this dual-labeling is useful in the training of our ACE-GAN since the samples generated by $G$ in the very early stage of training are far from satisfactory in mimicking the different beat types. Hence, we designed the $D$ to discriminate different beat types only when $D$ has determined it as real beat, which in turn regulates the learning of $G$. Actually, a similar approach (Mariani, Scheidegger, Istrate, Bekas, & Malossi, 2018) has been introduced to train GAN with imbalanced datasets. In addition, we used Mean Square Error (MSE) loss for the objective function of $\mathcal{L}_S$ instead as inspired by LSGAN (Mao et al., 2017) and Wasserstein distance (Arjovsky, Chintala, & Bottou, 2017; Gulrajani, Ahmed, Arjovsky, Dumoulin, & Courville, 2017). This added a penalty to the loss function to describe the distance of distribution between $\mathcal{X}_{fake}$ and $\mathcal{X}_{real}$. It also helps stabilize the GAN training process and assist in avoiding mode collapse problems (Mao et al., 2017).

The objective functions of our proposed ACE-GAN are now modified as follow:

$$\mathcal{L}_S = E[\text{MSE}\{S, s_{real}\} \mid \mathcal{X}_{real}] + E[\text{MSE}\{S, s_{fake}\} \mid \mathcal{X}_{fake}] \tag{5}$$

$$\mathcal{L}_C = E[\log P(C = c \mid \mathcal{X}_{real})] + E[\log P(C = c_{fake} \mid \mathcal{X}_{fake})] \tag{6}$$

where $D$ tries to maximize $(\mathcal{L}_C - \mathcal{L}_S)$ while $G$ learns to maximize $(\mathcal{L}_C + \mathcal{L}_S)$.

### 4.4. Implementation

Both the $G$ and $D$ in our ACE-GAN are randomly initialized from $\sim \mathcal{N}(0, 0.01)$. We used Adam optimizer (Kingma & Ba, 2015) with a learning rate of 0.0002 for both $G$ and $D$. Our proposed ACE-GAN is trained for a total of 10000 iterations. In each iteration, $D$ is updated twice (one with real data and one with generated data). $G$ is updated twice with the same batch of data to implement a two time-scale update rule (Heusel, Ramsauer, Unterthiner, Nessler, & Hochreiter, 2017), which helps in local convergence. We observed that this updating rule produces more stable trainings with smoother convergence in the losses in our GAN. The batch size is set to 128. In the training phase of $G$, a batch of $128 \times 1$ randomly assigned labels (including N, S, V, and F) and a $128 \times 100$ noise matrix are fed to $G$ twice, learning to generate coupling matrix samples ($\mathcal{X}_{fake}$). In the training phase of $D$, 32 samples from each of the 4 classes (N/S/V/F) are selected randomly from the common pool dataset, resulting in a batch of $32 \times 4 = 128$ training samples ($\mathcal{X}_{real}$) in total. $\mathcal{X}_{fake}$ is also fed to $D$ with "generated" class label. Through the gradients from $D$ in backpropagation, $G$ learns to generate new coupling matrices from the sampled noise according to the assigned labels. Detailed operating environment and hyper-parameters are shown in Table 3.

**Table 3**
Operating environment and hyper-parameters.

| | | | |
|---|---|---|---|
| Language | Python 3.6.8 | Dropout rate | 0.5 |
| Toolbox | Keras 2.2.4 (TensorFlow 1.12.0 backend) | BatchNorm momentum | 0.8 |
| CUDA version | 9.2 | Batch size | 128 |
| CUDNN version | 7.5 | Optimizer | Adam |
| Weight initialization | $\mathcal{N}(0,0.01)$ | Learning rate | $\alpha = 0.0002$, $\beta_1 = 0.5$, $\beta_2 = 0.999$ |

The well-trained $D$ is then extracted for classification purpose after further fine-tuning (Fig. 1). For each testing subject, the fine-tune dataset contains subject's own (estimated) N beats. Hence, the trained $D$ can be turned into patient-dependent classifiers for improved performance. The same optimizer and batch size in Table 3 are used for fine-tuning. The classifier is trained with the fine-tune dataset until the classification accuracy has reached 99%, or the absolute change of the accuracy is less than 1% in the last 10 epochs. Then the classifier is used to predict heartbeat labels for each record in DS2.

## 5. Evaluation

We here evaluated the classification performance for S and V beats detection as in previous studies (Acharya et al., 2017; Chazal, 2014; Chazal et al., 2004; Jiang & Kong, 2007; Kiranyaz et al., 2016; F. Li et al., 2019; Llamedo & Martinez, 2012; Mar et al., 2011; Rahhal et al., 2016; Takalo-Mattila, Kiljander, & Soininen, 2018; Zhai & Tin, 2018; Zhang et al., 2014). The following five metrics are calculated using the entire 30 min. of the records in DS2, including classification accuracy (Acc), sensitivity (Sen), specificity (Spe), positive predictive rate (Ppr) and $F_1$ score ($F_1$):

$$Acc = \frac{TP+TN}{TP+TN+FP+FN} \tag{7}$$

$$Sen = \frac{TP}{TP+FN} \tag{8}$$

$$Spe = \frac{TN}{TN+FP} \tag{9}$$

$$Ppr = \frac{TP}{TP+FP} \tag{10}$$

$$F_1 = 2 \cdot \frac{Sen \cdot Ppr}{Sen + Ppr} \tag{11}$$

where TP is the true positive, TN is the true negative, FP is the false positive and FN is the false negative.

We also evaluated our generated samples by Fréchet distance (FD), which has been widely used to assess GAN performance (Heusel et al., 2017),

$$FD = \|\mu_{real} - \mu_{generated}\|^2 + Tr\left(\Sigma_{real} + \Sigma_{generated} - 2\sqrt{\Sigma_{real}\Sigma_{generated}}\right) \tag{12}$$

where $\mu$ and $\Sigma$ are, respectively, the mean and covariance of the features from either real or generated samples. These features are obtained by feeding coupling matrices to the trained $D$ and extracting the outputs of its second last layer (150-dimensional vectors, Fig. 2b).

## 6. Results

We first show the effectiveness of GAN training in Fig. 3. Fig. 3a shows the adversarial losses for both $G$ and $D$ during the GAN training. MSE of $G$ becomes quite stable after ~2000 iterations of training, while MSE of $D$ decreases gradually, showing smaller loss values than $G$ in this minimax game. We also assessed the quality of generated samples by $G$ using FD (Eq. 12). For every 100 iterations, we calculated the FD between the generated samples and real samples (Fig. 3b). It shows that the FD also decreases gradually and converges after

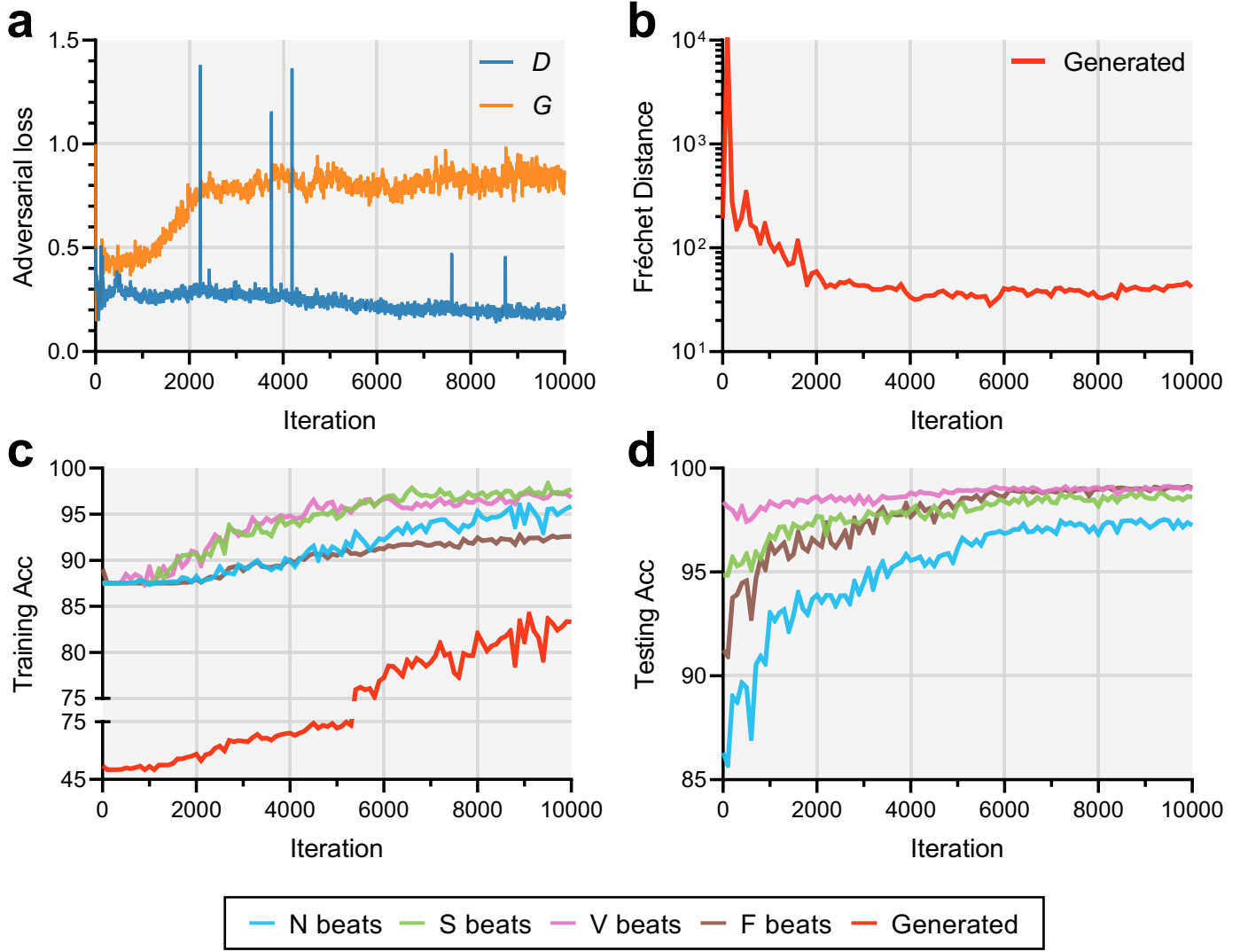

**Fig. 3.** Evaluation of the generator and the discriminator during the training with common pool dataset. (a) Adversarial MSE losses for the generator (*G*, orange) and the discriminator (*D*, blue) through a total of 10000 iterations. (b) Fréchet distance between generated samples and real beat samples through training. (c) Training classification accuracies of the discriminator tested on the 400 N/S/V/F real beats from common pool dataset and newly generated beats (400 per class). (d) Testing classification accuracies of the fine-tuned discriminator tested on DS2 dataset upon iterations of training.

~4000 iterations, indicating improved qualities of generated samples through training. We also tested the performance of *D* by its classification accuracy on both training (Fig. 3c) and testing (Fig. 3d) data. For training Acc, 1600 samples (400 per class) were selected randomly from the common pool dataset as well as generated by *G* respectively. The Acc for all of the four classes (N/S/V/F) of real beats gradually increased throughout training, which exceeded 90% eventually at 10000 iterations. (Fig. 3c). Acc for generated beats from *G* also increased gradually to over 80%. For testing Acc, upon every 100 iterations, we extracted the *D* and fine-tuned it with the fine-tune dataset (section 3.4) in the same way as the full system (section 4.4). The testing Acc for all four classes of beats increased gradually to over 95% at the end of simulation (Fig. 3d). Together, these imply that our GAN has been effectively trained, where both *G* and *D* serve well in generating and classifying samples, respectively.

**Table 4**
Confusion matrix for the extracted $D$ testing on DS2 without fine-tuning.

| | | Predicted labels | | | | | | Total |
|---|---|---|---|---|---|---|---|---|
| | | N | S | V | F | Q | Generated | |
| Ground truth labels | N | 20180 | 533 | 33 | 676 | 0 | 22837 | 44259 |
| | S | 207 | 375 | 15 | 0 | 0 | 1240 | 1837 |
| | V | 228 | 8 | 1370 | 9 | 0 | 1606 | 3221 |
| | F | 53 | 0 | 11 | 5 | 0 | 319 | 388 |
| | Q | 2 | 0 | 0 | 0 | 0 | 5 | 7 |
| Total | | 20670 | 916 | 1429 | 690 | 0 | 26007 | 49712 |

**Table 5**
Confusion matrix for the proposed ACE-GAN based fully automatic system testing on DS2.

| | | Predicted labels | | | | | | Total |
|---|---|---|---|---|---|---|---|---|
| | | N | S | V | F | Q | Generated | |
| Ground truth labels | N | 43593 | 266 | 142 | 87 | 0 | 171 | 44259 |
| | S | 145 | 1637 | 49 | 0 | 0 | 6 | 1837 |
| | V | 122 | 33 | 3019 | 45 | 0 | 2 | 3221 |
| | F | 293 | 0 | 34 | 61 | 0 | 0 | 388 |
| | Q | 2 | 0 | 3 | 1 | 0 | 1 | 7 |
| Total | | 44155 | 1936 | 3247 | 194 | 0 | 180 | 49712 |

Nevertheless, as shown in the confusion matrix in Table 4, the extracted $D$ would assign "generated" labels for a portion of real heartbeats when tested directly on DS2. To some extent, this shows that the extracted $D$ without fine-tuning could marginally discriminate between generated samples and un-seen real heartbeats not found in the common pool dataset. On the other hand, this also implies that our trained $G$ could generate heartbeat samples similar to the real heartbeats. Fig. 4a-4d present some examples of generated coupling matrices from different beat types alongside their real counterparts with similar patterns. For instance, the generated coupling matrix for N beat has a regular rhythm like the real one but with different R-R intervals and baseline levels (Fig. 4a). The generated matrices for abnormal beats show irregular rhythm similar to their real counterparts but with some variations (Fig. 4b-4d). Hence, our trained $G$ is not merely making replicates of heartbeats from a known record in a single patient but is able to generate meaningful samples with variation from the training data which may mimic the inputs for unseen patients. We further quantified the similarity between the real samples from the common pool dataset and the generated samples. We performed a principal component analysis (PCA) for features extracted from real/generated samples using the outputs of the second last layer in $D$ (see section 5). Fig. 4e shows the two PCs with the highest variances for features extracted from real/generated samples respectively. It is shown that the generated samples share similar distribution with the real samples of the same class in general. This validates the good quality of generated samples from our trained $G$. Meanwhile, simple decision boundaries could be sought among the four classes by applying linear discriminant analysis (LDA), indicating an effective feature extraction for the arrhythmia detection task with our proposed system. In fact, it is clear that the cluster of S beats is close to that of N beats. This highlights the difficulty in SVEB detection and also explains the need of S beat selection for training as discussed in Section 3.3. Furthermore, we also calculated the FD from generated samples to the real samples of different classes (Fig. 4f). We used all real samples from both DS1 and DS2, and generated the same amounts of fake samples as the real ones for each class. The FDs between real and generated samples of the same class are small (e.g. 42 for N beats and 58 for S beats), but those between samples of different classes are generally larger. These results again support that the trained $G$ can generate meaningful beats of different classes as in the real samples.

After GAN training and fine-tuning (section 4.4), the fine-tuned $D$ is used as the final arrhythmia classifier and tested on each subject in DS2. Table 5 shows the representative confusion matrix, where the mislabeled "generated" class is much reduced compared to Table 4. However, a small number of beats are still considered as "generated" by the classifier, probably due to their low fidelity or unprecedented abnormality. These beats are regarded as misclassification in our analysis.

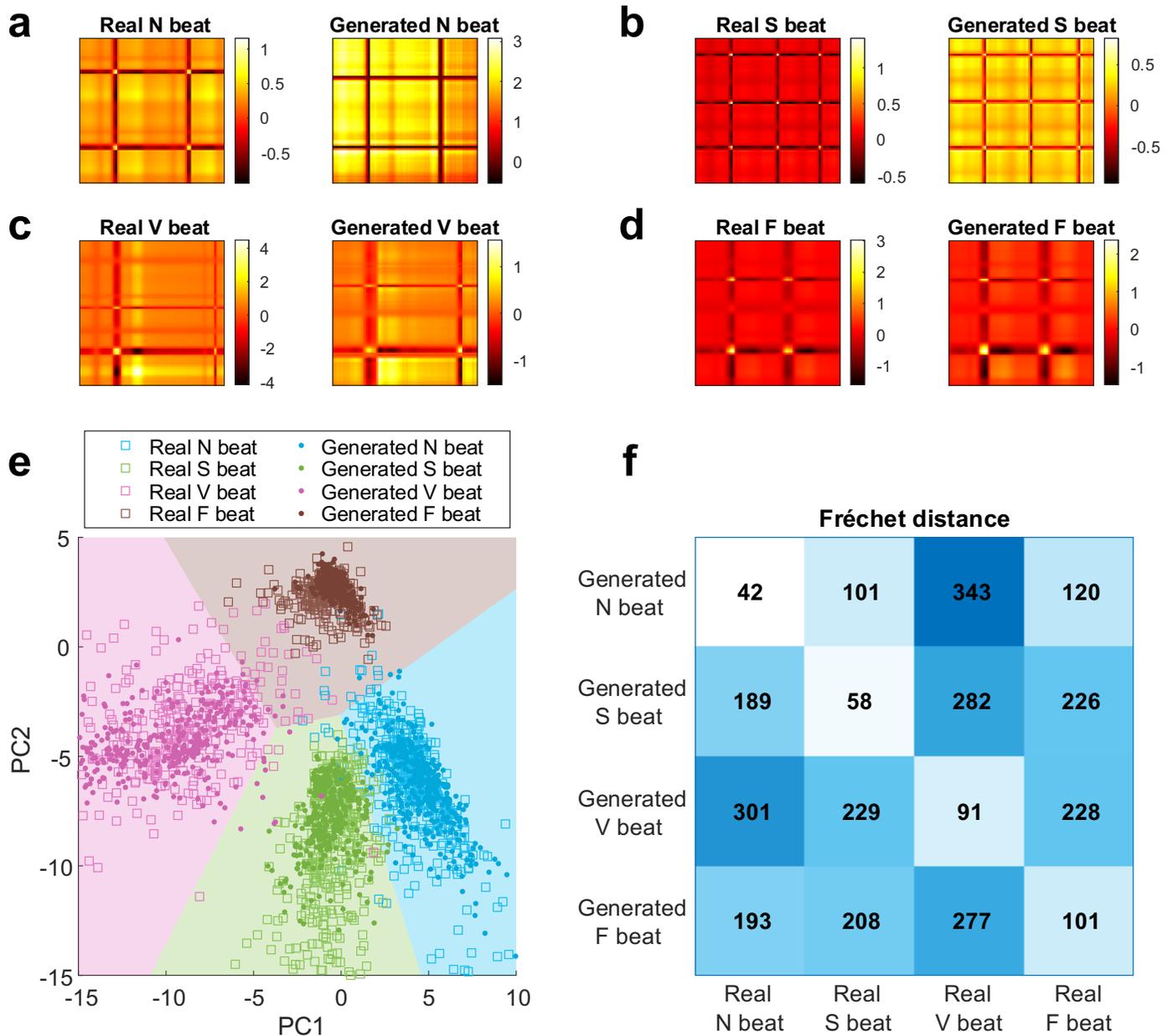

**Fig. 4.** Visual and statistical evaluation of the generated samples from the trained generator. (a) Representative example of real coupling matrix of N beat and its comparable generated counterpart. (b-d) Same as (a), but for S/V/F beat, respectively. (e) Principal component analysis of the features extracted from both real (400 per class from the common pool dataset) and generated (400 per class) beats. The decision boundaries for different classes are obtained by linear discriminant analysis. (f) Confusion matrix of Fréchet distances among real samples (all beats from DS1 and DS2) and generated samples (same number as all real beats per class) of various beat types.

Table 6 shows the comparison of the classification performance of our ACE-GAN with other state-of-art systems. Our simulations are repeated 10 times where $G$ generates different random input samples. The performance of our classifiers remains quite stable with small variations. Our fully automatic system achieves very high Acc and Spe for both SVEB (99% and 99% respectively) and VEB (99% and 99% respectively), which are comparable with several state-of-art methods (Table 6). More importantly, the Sen for SVEB (87%) and VEB (93%) in our system are higher than those of all previous automatic systems in Table 6 (Chazal et al., 2004; Garcia, Moreira, Menotti, & Luz, 2017; Llamedo & Martinez, 2011, 2012; Mar et al., 2011; Raj & Ray, 2018; Ye et al., 2016; Zhang et al., 2014). In fact, the Sen for SVEB and VEB of our system are comparable with some of the expert assisted systems, which require manual annotation for up to 500 beats or 5 minutes of ECG recording for each testing subject. Such high Sen is important for practical clinical use. Furthermore, our Ppr of SVEB indeed ranks second among all these automatic methods compared, while Ppr of VEB is quite high (94%). Consequently, relatively high $F_1$ scores for both VEB (93%) and SVEB (86%) are achieved. Actually, our $F_1$ score for SVEB is higher than the top-ranked automatic method in Table 6 by up to 10%.

**Table 6**
SVEB and VEB classification performance of the proposed systems and comparison with former studies (22 testing records in DS2).

| | Fully automatic | # labels required / patient | SVEB (%) | | | | | VEB (%) | | | | |
|---|---|---|---|---|---|---|---|---|---|---|---|---|
| | | | Acc | Sen | Spe | Ppr | F₁ | Acc | Sen | Spe | Ppr | F₁ |
| **Automatic systems** | | | | | | | | | | | | |
| Mar et al., 2011 | | | 93.3 | 83.2 | 93.7 | 33.5 | 47.8 | 97.4 | 86.8 | 98.1 | 75.9 | 81.0 |
| Zhang et al., 2014 | | | 93.3 | 79.1 | 93.9 | 36.0 | 49.5 | 98.6 | 85.5 | 99.5 | 92.7 | 89.0 |
| Chazal et al., 2004 | | | 94.6 | 75.9 | 95.4 | 38.5 | 55.1 | 97.4 | 77.7 | 98.8 | 81.9 | 79.7 |
| Llamedo & Martínez, 2011 | | | 95.1 | 76.5 | 95.8 | 41.3 | 53.6 | 98.2 | 83.1 | 99.2 | 88.1 | 85.5 |
| Garcia et al., 2017 | | | 96.6 | 62.0 | 97.9 | 53.0 | 57.1 | 95.4 | 87.3 | 95.9 | 59.4 | 70.7 |
| Llamedo & Martínez, 2012 | Yes | - | - | 79 ± 2 | - | 46 ± 2 | 58.1 | - | 89 ± 1 | - | 87 ± 1 | 88.0 |
| Takalo-Mattila et al., 2018 | | | - | 62.0 | - | 56.0 | 58.8 | - | 89.0 | - | 51.0 | 64.8 |
| Raj & Ray, 2018 | | | 96.2 | 80.8 | 96.7 | 48.8 | 60.8 | 97.9 | 82.2 | 99.0 | 85.4 | 83.8 |
| Ye et al., 2016 | | | 98.3 | 61.4 | 99.8 | 90.7 | 73.2 | 99.4 | 91.8 | 99.9 | 98.3 | 94.9 |
| Niu et al., 2020 | | | - | 76.5 | 99.1 | 76.6 | 76.6 | - | 85.7 | 99.6 | 94.1 | 89.7 |
| **Proposed ACE-GAN** | | | **99 ± 0** | **87 ± 2** | **99 ± 0** | **85 ± 1** | **86 ± 1** | **99 ± 0** | **93 ± 1** | **99 ± 0** | **94 ± 1** | **93 ± 0** |
| **Expert assisted systems** | | | | | | | | | | | | |
| Chazal & Reilly, 2006 | | 500 beats | 95.8 | 87.7 | 96.1 | 47.0 | 61.2 | 99.4 | 94.3 | 99.7 | 96.2 | 95.2 |
| Llamedo & Martínez, 2012 | | 1 beat | - | 83 ± 4 | - | 58 ± 5 | 68.3 | - | 91 ± 3 | - | 90 ± 2 | 90.5 |
| Chazal, 2014 | No | 100 beats | 97.8 | 94.0 | - | 62.5 | 75.1 | 99.4 | 93.4 | - | 97.0 | 95.2 |
| Li et al. 2019 | | 5 min. | - | 86.2 | - | 72.5 | 78.8 | - | 92.2 | - | 88.5 | 90.3 |
| Ye et al., 2016 | | 5 min. | 99.1 | 76.5 | 99.9 | 99.1 | 86.3 | 99.7 | 97.1 | 99.9 | 98.5 | 97.8 |
| Llamedo & Martínez, 2012 | | 12 beats | - | 92 ± 1 | - | 90 ± 3 | 91.0 | - | 93 ± 1 | - | 97 ± 1 | 95.0 |

**Table 7**
Detailed classification performance of 22 testing records in DS2.

| Record | Number of beats | | | N (%) | | SVEB (%) | | VEB (%) | | Overall |
|---|---|---|---|---|---|---|---|---|---|---|
| | N | SVEB | VEB | Sen | Ppr | Sen | Ppr | Sen | Ppr | Acc |
| 100 | 2239 | 33 | 1 | 100 ± 0 | 100 ± 0 | 100 ± 0 | 97 ± 0 | 100 ± 0 | 100 ± 0 | 100 ± 0 |
| 103 | 2082 | 2 | 0 | 100 ± 0 | 100 ± 0 | 100 ± 0 | 52 ± 10 | - | - | 100 ± 0 |
| 105 | 2526 | 0 | 41 | 96 ± 0 | 100 ± 0 | - | - | 90 ± 2 | 50 ± 5 | 95 ± 0 |
| 111 | 2123 | 0 | 1 | 99 ± 0 | 100 ± 0 | - | - | 100 ± 0 | 17 ± 13 | 99 ± 0 |
| 113 | 1789 | 6 | 0 | 99 ± 0 | 100 ± 0 | 23 ± 15 | 29 ± 20 | - | - | 99 ± 0 |
| 117 | 1534 | 1 | 0 | 100 ± 0 | 100 ± 0 | 0 ± 0 | 0 ± 0 | - | - | 100 ± 0 |
| 121 | 1861 | 1 | 1 | 100 ± 0 | 100 ± 0 | 90 ± 30 | 43 ± 31 | 100 ± 0 | 55 ± 24 | 100 ± 0 |
| 123 | 1515 | 0 | 3 | 99 ± 0 | 100 ± 0 | - | - | 23 ± 15 | 0 ± 0 | 99 ± 0 |
| 200 | 1743 | 30 | 826 | 99 ± 0 | 99 ± 0 | 67 ± 8 | 42 ± 3 | 97 ± 0 | 100 ± 0 | 98 ± 0 |
| 202 | 2061 | 55 | 19 | 99 ± 0 | 99 ± 0 | 47 ± 5 | 59 ± 5 | 95 ± 0 | 66 ± 8 | 98 ± 0 |
| 210 | 2423 | 22 | 195 | 99 ± 0 | 98 ± 0 | 16 ± 8 | 36 ± 7 | 91 ± 2 | 94 ± 2 | 98 ± 0 |
| 212 | 2748 | 0 | 0 | 100 ± 0 | 100 ± 0 | - | - | - | - | 100 ± 0 |
| 213 | 2641 | 28 | 220 | 99 ± 1 | 90 ± 0 | 22 ± 11 | 89 ± 8 | 85 ± 6 | 76 ± 3 | 88 ± 0 |
| 214 | 2003 | 0 | 256 | 99 ± 1 | 99 ± 0 | - | - | 92 ± 3 | 97 ± 0 | 98 ± 0 |
| 219 | 2082 | 7 | 64 | 98 ± 0 | 99 ± 0 | 1 ± 4 | 1 ± 2 | 71 ± 9 | 73 ± 5 | 97 ± 0 |
| 221 | 2031 | 0 | 396 | 99 ± 0 | 100 ± 0 | - | - | 100 ± 0 | 100 ± 0 | 100 ± 0 |
| 222 | 2274 | 209 | 0 | 92 ± 2 | 97 ± 1 | 68 ± 6 | 53 ± 3 | - | - | 90 ± 1 |
| 228 | 1688 | 3 | 362 | 98 ± 0 | 100 ± 0 | 67 ± 0 | 37 ± 5 | 98 ± 0 | 99 ± 0 | 98 ± 0 |
| 231 | 1568 | 1 | 2 | 97 ± 1 | 100 ± 0 | 0 ± 0 | 0 ± 0 | 90 ± 20 | 26 ± 10 | 97 ± 1 |
| 232 | 398 | 1382 | 0 | 100 ± 0 | 92 ± 4 | 97 ± 2 | 100 ± 0 | - | - | 98 ± 1 |
| 233 | 2230 | 7 | 831 | 99 ± 0 | 97 ± 1 | 9 ± 7 | 0 ± 0 | 90 ± 1 | 98 ± 0 | 96 ± 0 |
| 234 | 2700 | 50 | 3 | 100 ± 0 | 99 ± 0 | 42 ± 4 | 100 ± 0 | 100 ± 0 | 53 ± 17 | 99 ± 0 |
| **Total** | **44259** | **1837** | **3221** | **99 ± 0** | **99 ± 0** | **87 ± 2** | **85 ± 1** | **93 ± 1** | **94 ± 1** | **97 ± 0** |

Table 7 shows the performance of our classifiers for each subject in DS2. It shows that almost all subjects had very high Sen and Ppr (>95%) for N beats prediction. Sen of VEB is generally high for most subjects. However, Sen of SVEB varies quite significantly from subject to subject. Ppr is inevitably low in those subjects with small number of SVEB or VEB. Nevertheless, for subjects who have over 1% SVEB in their own record, an average Sen of 70% and an average Ppr of 75% for SVEB detection can still be achieved. As such, the classification system can still provide reasonable performance for these challenging cases.

## 7. Discussion

In this study, we have proposed an automatic ECG arrhythmia classifier ($D$) based on a GAN framework to achieve desirable performance.

To examine how the $D$ contributes to ECG arrhythmia classifications, we have performed further simulations. Fig. 5 shows the performance of our system with or without using the extracted $D$, when fine-tuned with different numbers of generated samples from $G$, respectively. In the case without the extracted $D$, we used a classifier with the same structure as $D$ but with randomly initialized weights. We refer to such classifier as the "fresh classifier" in the following discussion, since it is fresh from the information learnt by the GAN. The "fresh classifier" is trained on the same fine-tune dataset (see Section 3.4) as we fine-tune the extracted $D$. All training conditions for both classifiers remain the same. Fig. 5 illustrates that using the extracted $D$ can boost the overall classification performance by over 15% for the $F_1$ of SVEB (Fig. 5a) and around 5% for that of VEB (Fig. 5d). In particular, the Ppr of SVEB is improved dramatically by over 25% (Fig. 5c). For VEB, a 5% increase can also be observed for Sen (Fig. 5e) and 10% for Ppr (Fig. 5f). Therefore, it suggests that the extracted $D$ possesses richer information favorable to arrhythmia detection than the "fresh classifier", which could come from its interplay with numerous generated samples from $G$ and the whole common pool dataset throughout the training. The "fresh classifier" has higher Sen of SVEB than the one using trained weights from $D$ when no generated beats are included (Fig. 5b). However, it comes at a cost of very low Ppr of SVEB (<45%, Fig. 5c) and consequently, low $F_1$ of SVEB (<60%, Fig. 5a) which is inappropriate for real-world application.

Fig. 5 also shows the effect of different numbers of generated samples per class on the classification performance. Generally speaking, the generated beats mildly improve the performance (by ~5%) in the case of using extracted $D$. However, the Sen increases at the cost of declining Ppr. A compromise between Sen and Ppr, hence, needs to be considered. The effects of generated samples are much larger in the case of using the "fresh classifier". On one hand, it implies that the trained $D$ has already possessed significant information about distribution of generated samples from $G$, as compared to a "fresh" network. Hence, providing more generated samples to the trained $D$ has minor effect to improve its classification performance. This again demonstrates that the trained $G$ could generate informative samples to help in classification tasks. On the other hand, it is possible that the trained $G$ in GAN generates some distorted samples occasionally. However, these distorted samples are small in number whose influence is mild as physiological signals are intrinsically noisy, and this can be alleviated by providing more generated samples. Essentially, the construction of our 2-D coupling matrix input using a pair of adjacent heartbeat segments could be regarded as a $2^{nd}$ order polynomial feature transform, which is a simple and commonly used approach for seeking linearly separable features in higher-dimensional space. However, our coupling matrix is also geometrically structured to preserve physiological information, which is preferable for CNN to extract features automatically.

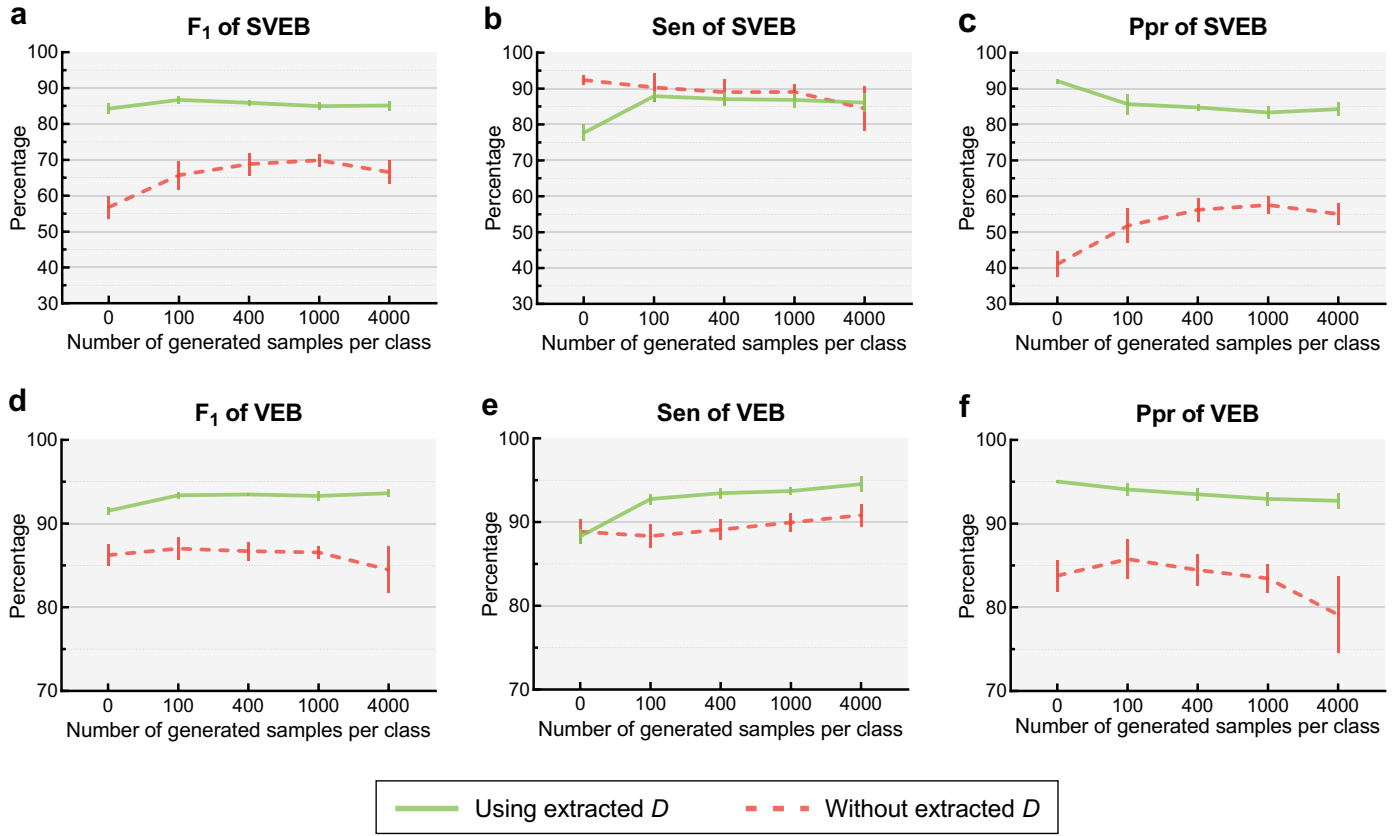

**Fig. 5.** Comparison of performance with or without using the weights from the trained $D$, when including different numbers of generated samples from $G$ in the fine-tune dataset. Green: the trained $D$ is extracted to initialize the ECG classifier; Red: the ECG classifier is randomly initialized. Error bars show standard derivation. All classifiers are fine-tuned with the fine-tune dataset, which includes the subset of common pool dataset, the automatically estimated N beats, and 0-4000 generated samples per class.

In our system, the fine-tuning step serves three purposes. First, as illustrated in Section 6 (Tables 4-5), the extracted $D$ without any fine-tuning labeled a number of real beats as "generated" due to the design of the GAN training framework. Hence, during the fine-tuning step, all training inputs are considered as "real" samples, in order to push the final classifier to predict the type of beats, regardless of the source (real or generated). We show that this approach is effective and yields promising results (Tables 5-7). Second, fine-tuning with additional beats from $G$ serves a data augmentation purpose to improve the performance (Fig. 5). However, apparently, the role of $G$ for data augmentation seems to contribute more during the training of the GAN. The iterative training of $G$ and $D$ has fed $D$ with numerous generated samples with progressively improving quality. In other words, the $G$ has contributed to augment data in both the training step (major) and the fine-tuning step (minor). Third, fine-tuning with the subject-specific estimated N beats turns the classifier into a patient-dependent classifier by providing subject-specific information. It significantly increases the Ppr of SVEB (58% to 85%), and consequently, the corresponding $F_1$ score has also been improved from 69% to 86% (Tables 6-8). The performance of VEB detection also improves slightly. This indicates that these estimated N beats are useful for arrhythmia detection. It is shown that

**Table 8**
Performance of the proposed classification system with different components of fine-tune dataset.

| | Fine-tune dataset | | | SVEB (%) | | | | | VEB (%) | | | | |
|---|---|---|---|---|---|---|---|---|---|---|---|---|---|
| | Estimated N beats | Sub-set of common pool | Generated samples | Acc | Sen | Spe | Ppr | $F_1$ | Acc | Sen | Spe | Ppr | $F_1$ |
| a) | √ | √ | × | 99 ± 0 | 78 ± 2 | 99 ± 0 | 92 ± 1 | 84 ± 1 | 99 ± 0 | 88 ± 1 | 99 ± 0 | 95 ± 0 | 92 ± 0 |
| b) | √ | × | √ | 99 ± 0 | 89 ± 1 | 99 ± 0 | 80 ± 1 | 85 ± 1 | 99 ± 0 | 94 ± 1 | 99 ± 0 | 92 ± 1 | 93 ± 1 |
| c) | × | √ | √ | 97 ± 0 | 87 ± 2 | 98 ± 0 | 58 ± 2 | 69 ± 1 | 99 ± 0 | 94 ± 1 | 99 ± 0 | 87 ± 4 | 90 ± 2 |

a) Equivalent to Fig. 5 using extracted *D* with 0 generated samples.
b) 400 S/V/F beats per class are generated.
c) The estimated N beats are replaced with 400 N beats selected randomly from the common pool dataset.

**Table 9**
Accuracy of N beats estimation from 22 testing records in DS2.

| Record | Correct number | Incorrect number | Accuracy | Record | Correct number | Incorrect number | Accuracy |
|---|---|---|---|---|---|---|---|
| 100 | 400 | 0 | 100.0% | 213 | 387 | 13 | 96.8% |
| 103 | 400 | 0 | 100.0% | 214 | 400 | 0 | 100.0% |
| 105 | 400 | 0 | 100.0% | 219 | 395 | 5 | 98.8% |
| 111 | 400 | 0 | 100.0% | 221 | 400 | 0 | 100.0% |
| 113 | 400 | 0 | 100.0% | 222 | 231 | 0 | 100.0% |
| 117 | 400 | 0 | 100.0% | 228 | 106 | 0 | 100.0% |
| 121 | 400 | 0 | 100.0% | 231 | 400 | 0 | 100.0% |
| 123 | 400 | 0 | 100.0% | 232 | - | - | - |
| 200 | 29 | 0 | 100.0% | 233 | 400 | 0 | 100.0% |
| 202 | 400 | 0 | 100.0% | 234 | 400 | 0 | 100.0% |
| 210 | 400 | 0 | 100.0% | | | | |
| 212 | 400 | 0 | 100.0% | Avg. | 343 | 1 | 99.7% |

combining the common pool, the generated samples, and the estimated N beats achieves the best overall performance (Tables 6-8). Generally speaking, including generated samples help to boost the Sen while the estimated N beats contribute to maintaining high Ppr levels.

We would also like to point out that the subject-specific N beats can be reliably extracted in an unsupervised way, hence allowing the system to remain fully automatic, unlike those typical patient-dependent classifiers which require expert intervention for each patient. Table 9 presents the accuracy of the proposed N beats estimation method for each patient in DS2. Up to 400 N beats were estimated for each patient. The overall accuracy for estimating N beats is 99.7%, which is significantly higher than the chance level of 89.0% in DS2. Note that 100% accuracy has been achieved in 19 out of 22 patients in DS2. No N beat could be estimated from #232 because sinus bradycardia occupied the whole 30 min. recording and there were numerous long pauses up to 6 sec. such that regular rhythm was difficult to detect.

Table 6 indicates that our proposed system clearly has superior performance in SVEB prediction than all the automatic systems on the list, except for the Ppr in (Ye et al., 2016). Their automatic system achieved a very high Ppr of SVEB (90.7%) and VEB (98.3%). However, both their automatic system and expert assisted system suffered from low Sen of SVEB (61.4% and 76.5%), as shown in Table 6, which greatly limits the viability of practical clinical usage.

Data augmentation for time series signal is an active area of research (Iwana & Uchida, 2020). We herein compare our proposed automatic system with some recent studies that attempted to improve performance of arrhythmia or abnormal ECG detection by data augmentation (Table 10). Three recent studies (Golany & Radinsky, 2019; Shaker, Tantawi, Shedeed, & Tolba, 2020; P. Wang, Hou, Shao, & Yan, 2019) applied GAN based models to synthesize 1-D ECG signals for data augmentation and to provide a more balanced training set. The real and synthesized data were then used to train a DNN classifier. However, for (Shaker et al., 2020; P. Wang et al., 2019), the effectiveness of their generated samples in real world application remains unclear, since they mixed the beats of all subjects together, without following the clinical

**Table 10**
Comparison with related work using ECG data augmentation.

| Method | Proposed | Wang et al. | Shaker et al. | Golany et al. | Nonaka et al. | Nonaka et al. | Nankani et al. |
|---|---|---|---|---|---|---|---|
| Application | **Arrhythmia detection** | Arrhythmia detection | Arrhythmia detection | Arrhythmia detection | Arrhythmia detection | Abnormal ECG detection | - |
| Following clinical convention | **Yes** | No | No | Yes^ | No | - | - |
| Performance gain with data augmentation* | **∼ 25% in $F_1$** | ∼ 8% in Sen | ∼ 4% in Ppr | ∼ 20% in AUC | ∼ 3% in $F_1$ | - | - |
| Data augmentation framework | **ACE-GAN** | AC-GAN | GAN | PGAN | Random transformation | Random transformation | DCCGAN |
| Classification framework | | LSTM with ResNet | InceptionNet | LSTM | Deep CNN with ResNet | EfficientNet | - |
| Requiring separate classifier | **No (Unified)** | Yes | Yes | Yes | Yes | Yes | - |
| Format of augmented data | **2-D coupling matrix** | 1-D waveform | 1-D waveform | 1-D waveform & features | 1-D waveform | 1-D waveform | 1-D waveform |
| Class conditioned augmentation | **Yes** | Yes | No | No | - | - | Yes |
| Visual evaluation of generated data | **Yes** | Yes | Yes | No | No | No | No |
| Statistical evaluation of generated data | **Yes** | Yes | No | No | No | No | Yes |

- Not applicable or not reported.
^ Tested on subjects with over 800 arrhythmic beats only.
* Measured by increments between classification performance with/without the augmentation frameworks.

convention, where record of each testing subject should be considered separately (Chazal et al., 2004; Chazal & Reilly, 2006; Llamedo & Martinez, 2011, 2012; Mar et al., 2011; Raj & Ray, 2018; Teijeiro et al., 2018; Ye et al., 2016; Zhang et al., 2014). For (Golany & Radinsky, 2019), the effectiveness is also unclear since they only evaluated on subjects with abundant SVEB or VEB (over 800 arrhythmias in either class). Some studies (Nonaka & Seita, 2020a, 2020b) also attempted to augment ECG data with random transformation technique, but they used separate classifiers for downstream detection tasks. In (Nankani & Baruah, 2020), although the data generated by GAN has been assessed with multiple statistical metrics, the actual benefit in the arrhythmia detection was not studied. In this work, on the contrary, we proposed a unified GAN based framework for both augmenting class conditioned coupling matrices and detecting arrhythmia, where a separate classifier is not required. Following the clinical convention, a performance gain of up to 25% in $F_1$ score has been achieved (compared to the system without extracted $D$ and generated samples, Fig. 5a).

Our proposed system was tested on the MIT-BIH arrhythmia database (A. L. Goldberger et al., 2000), which was specially built for the evaluation of arrhythmia detectors (Moody & Mark, 2001). Historically, this is the sole dataset with a widely recognized data partition approach (i.e. separation of training set and testing set) that was considered by almost all arrhythmia classification studies for standardized benchmarking. Some researches (Llamedo & Martinez, 2011; Tan, Zhang, Wu, & Qian, 2019) also included the MIT-BIH supraventricular arrhythmia database (A. L. Goldberger et al., 2000) as a supplement for training or fine-tuning (since the number of S beats in MIT-BIH arrhythmia database is relatively limited), but not for the purpose of performance evaluation. In fact, we show in this study that the data imbalance problem could be readily settled with our proposed system. In future work, actual clinical routines with daily ECG records could further testify the performance and the automation of this system.

## 8. Conclusion

In this study, we have proposed a novel integrated framework for data augmentation and ECG beat classification based on ACE-GAN. Our system implements a GAN to synthesize additional arrhythmic heartbeats for data augmentation. After adversarial training, the discriminator is extracted and fine-tuned into a patient-dependent classifier. Its performance in detection for SVEB and VEB outperforms several state-of-art automatic systems and also some expert assisted methods. We have performed detailed simulations and analyses to present the contribution of each component in our system. This work reveals the effectiveness of GAN based systems in ECG classification, and the potential of our proposed system as a promising tool for clinical arrhythmic beats screening.

## Acknowledgment

This work was supported by Research Grants Council of Hong Kong SAR (Project CityU 11213717).